\documentclass[11pt,a4paper]{article}
\usepackage[hyperref]{acl2018}
\usepackage{times}
\usepackage{latexsym}
\usepackage{algorithm}
\usepackage{amsmath}
\usepackage{amssymb}
\usepackage[T1]{fontenc}
\usepackage{graphicx}
\usepackage{amsfonts}
\usepackage{booktabs}
\usepackage{subcaption}
\usepackage[]{algpseudocode}
\usepackage{amsthm}
\usepackage{relsize}
\usepackage{amsmath}
\usepackage{multirow}
\usepackage{url}
\usepackage{color,soul}
\aclfinalcopy 
\def\aclpaperid{977} 

\theoremstyle{definition}
\newtheorem{exmp}{Example}[section]

\theoremstyle{definition}
\newtheorem{definition}{Definition}[section]

\title{Straight to the Tree: Constituency Parsing \\
with Neural Syntactic Distance}
\author{
  Yikang Shen\thanks{\;\;Equal contribution. Corresponding authors: yi-kang.shen@umontreal.ca, zhouhan.lin@umontreal.ca.}\;\,\thanks{\;\;Work done while at Microsoft Research, Montreal.} \\ MILA \\
  University of Montr\'eal \\ \And
  Zhouhan Lin$^{*\dagger}$ \\
  MILA \\
  University of Montr\'eal \\
  AdeptMind Scholar \And
  Athul Paul Jacob$^{\dagger}$ \\
  MILA \\ 
  University of Waterloo \\\AND
  Alessandro Sordoni \\
  Microsoft Research \\
  Montr\'eal, Canada \\\And
  Aaron Courville \and Yoshua Bengio \\
  MILA \\
  University of Montr\'eal, CIFAR \\
  }

\date{}

\begin{document}
\maketitle
\begin{abstract}
In this work, we propose a novel constituency parsing scheme. The model predicts a vector of real-valued scalars, named syntactic distances, for each split position in the input sentence. The syntactic distances specify the order in which the split points will be selected, recursively partitioning the input, in a top-down fashion.
Compared to traditional shift-reduce parsing schemes, our approach is free from the potential problem of compounding errors, while being faster and easier to parallelize. Our model achieves competitive performance amongst single model, discriminative parsers in the PTB dataset and outperforms previous models in the CTB dataset.
\end{abstract}

\section{Introduction}

Devising fast and accurate constituency parsing algorithms is an important, long-standing problem in natural language processing. Parsing has been useful for incorporating linguistic prior in several related tasks, such as relation extraction, paraphrase
detection~\citep{callison2008syntactic}, and more recently, natural language inference~\citep{bowman2016fast} and machine translation~\cite{eriguchi2017learning}.

Neural network-based approaches relying on dense input representations have recently achieved competitive results for constituency parsing~\citep{,vinyals2015grammar,cross2016span,liu2016shift,stern2017minimal}. Generally speaking, either these approaches produce the parse tree sequentially, by governing the sequence of transitions in a transition-based parser~\citep{nivre2004incrementality,zhu2013fast,chen2014fast,cross2016span},
or use a chart-based approach by estimating non-linear potentials and performing exact structured inference by dynamic programming~\cite{finkel2008efficient,durrett2015neural,stern2017minimal}.

Transition-based models decompose the structured prediction problem into a sequence of local decisions. This enables fast greedy decoding but also leads to compounding errors because the model is never exposed to its own mistakes during training~\citep{daume2009search}. Solutions to this problem usually complexify the training procedure by using structured training through beam-search~\citep{weiss2015structured,andor2016globally} and dynamic oracles~\citep{goldberg2012dynamic,cross2016span}. On the other hand, chart-based models can incorporate structured loss functions during training and benefit from exact inference via the CYK algorithm but suffer from higher computational cost during decoding~\cite{durrett2015neural,stern2017minimal}.

\begin{figure}[!t]
  \centering
  \includegraphics[width=0.55\linewidth]{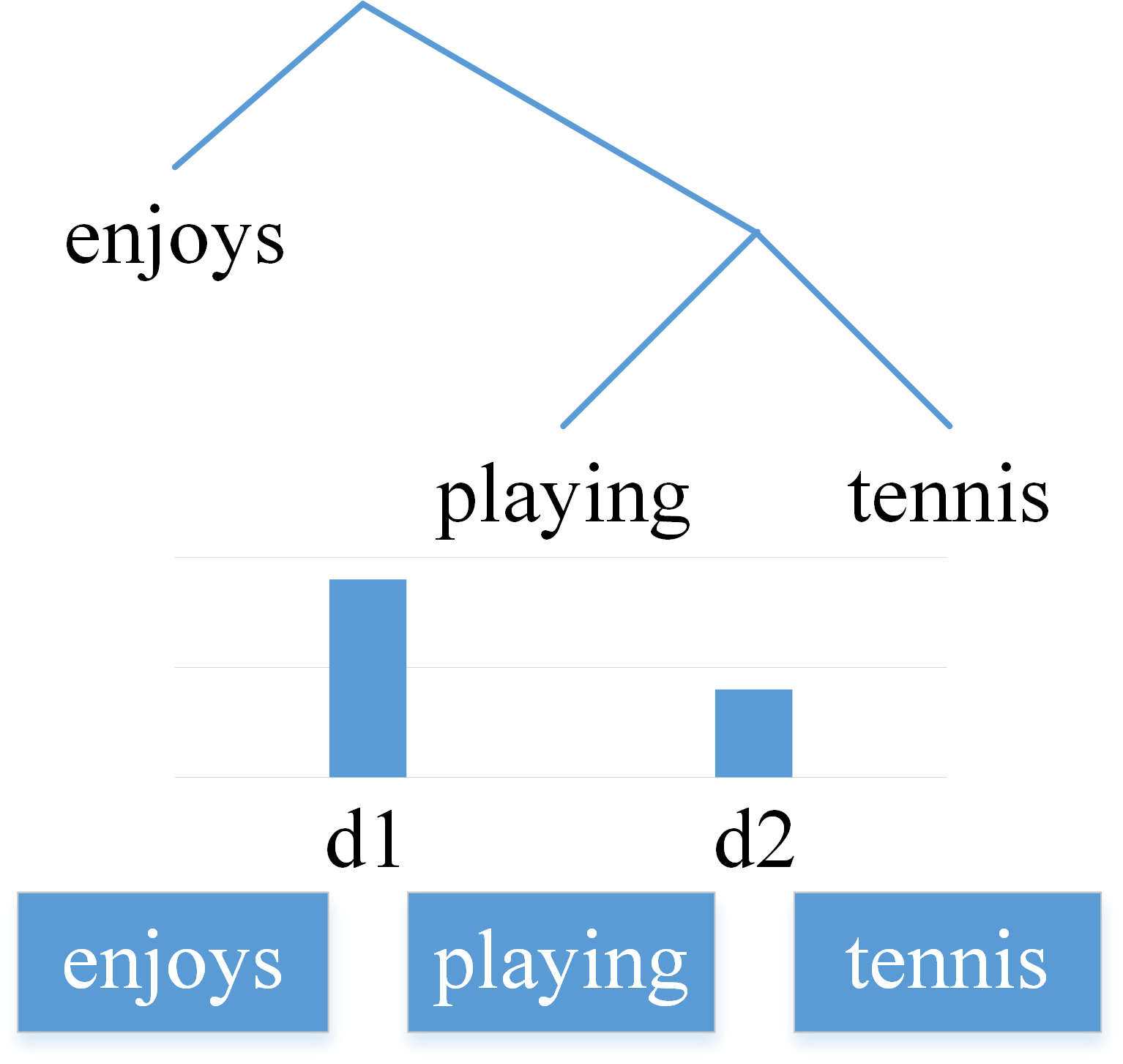}
  \caption{An example of how syntactic distances ($d1$ and $d2$) describe the structure of a parse tree: consecutive words with larger predicted distance are split earlier than those with smaller distances, in a process akin to divisive clustering.}
  \label{fig_dis2tree_example}
\end{figure}

In this paper, we propose a novel, fully-parallel model for constituency parsing, based on the concept of ``syntactic distance'', recently introduced by~\citep{shen2017neural} for language modeling. 
To construct a parse tree from a sentence, one can proceed in a top-down manner, recursively splitting larger constituents into smaller constituents, where the order of the splits defines the hierarchical structure.
The syntactic distances are defined for each possible split point in the sentence. The order induced by the syntactic distances fully specifies the order in which the sentence needs to be recursively split into smaller constituents (Figure \ref{fig_dis2tree_example}):
in case of a binary tree, there exists a one-to-one correspondence between the ordering and the tree.
Therefore, our model is trained to reproduce the ordering between split points induced by the ground-truth distances by means of a margin rank loss~\citep{weston2011wsabie}.
Crucially, our model works \emph{in parallel}: the estimated distance for each split point is produced independently from the others, which allows for an easy parallelization in modern parallel computing architectures for deep learning, such as GPUs. 
Along with the distances, we also train the model to produce the constituent labels, which are used to build the fully labeled tree.

Our model is fully parallel and thus does not require computationally expensive structured inference during training. Mapping from syntactic distances to a tree can be efficiently done in $\mathcal{O}(n \log n)$, which makes the decoding computationally attractive. Despite our strong conditional independence assumption on the output predictions, we achieve good performance for single model discriminative parsing in PTB (91.8 F1) and CTB (86.5 F1) matching, and sometimes outperforming, recent chart-based and transition-based parsing models.

\section{Syntactic Distances of a Parse Tree} \label{sec:dis_n_tree}

\begin{figure*}[!h]
	\centering
    \begin{subfigure}[b]{0.58\textwidth}
    	\centering
        \includegraphics[width=0.95\textwidth]{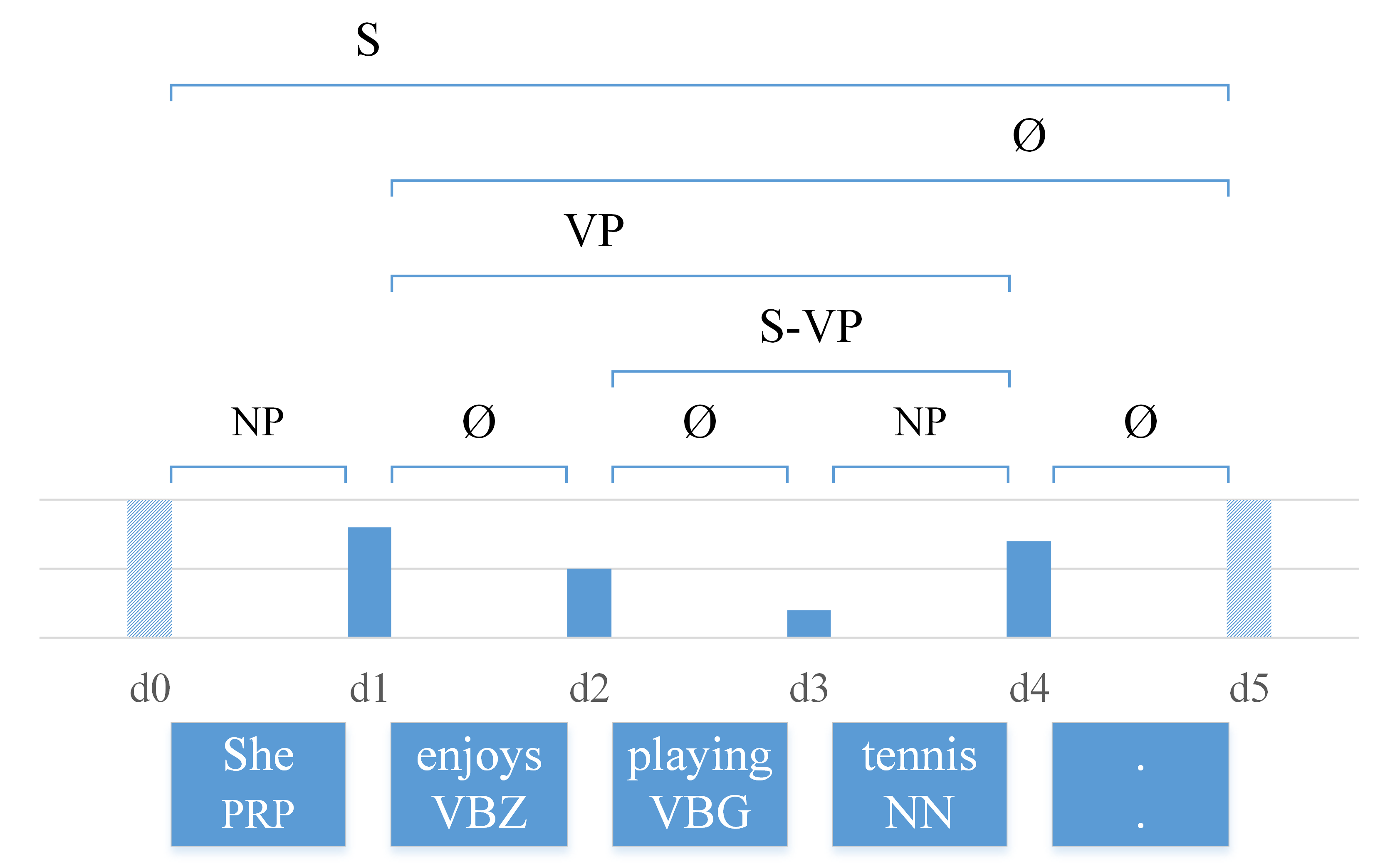}
        \caption{Boxes in the bottom are words and their corresponding POS tags predicted by an external tagger. The vertical bars in the middle are the syntactic distances, and the brackets on top of them are labels of constituents. The bottom brackets are the predicted unary label for each words, and the upper brackets are predicted labels for other constituent.}
        \label{fig_infer_tree}
    \end{subfigure}
    ~
    \begin{subfigure}[b]{0.4\textwidth}
    	\centering
        \includegraphics[width=0.8\textwidth]{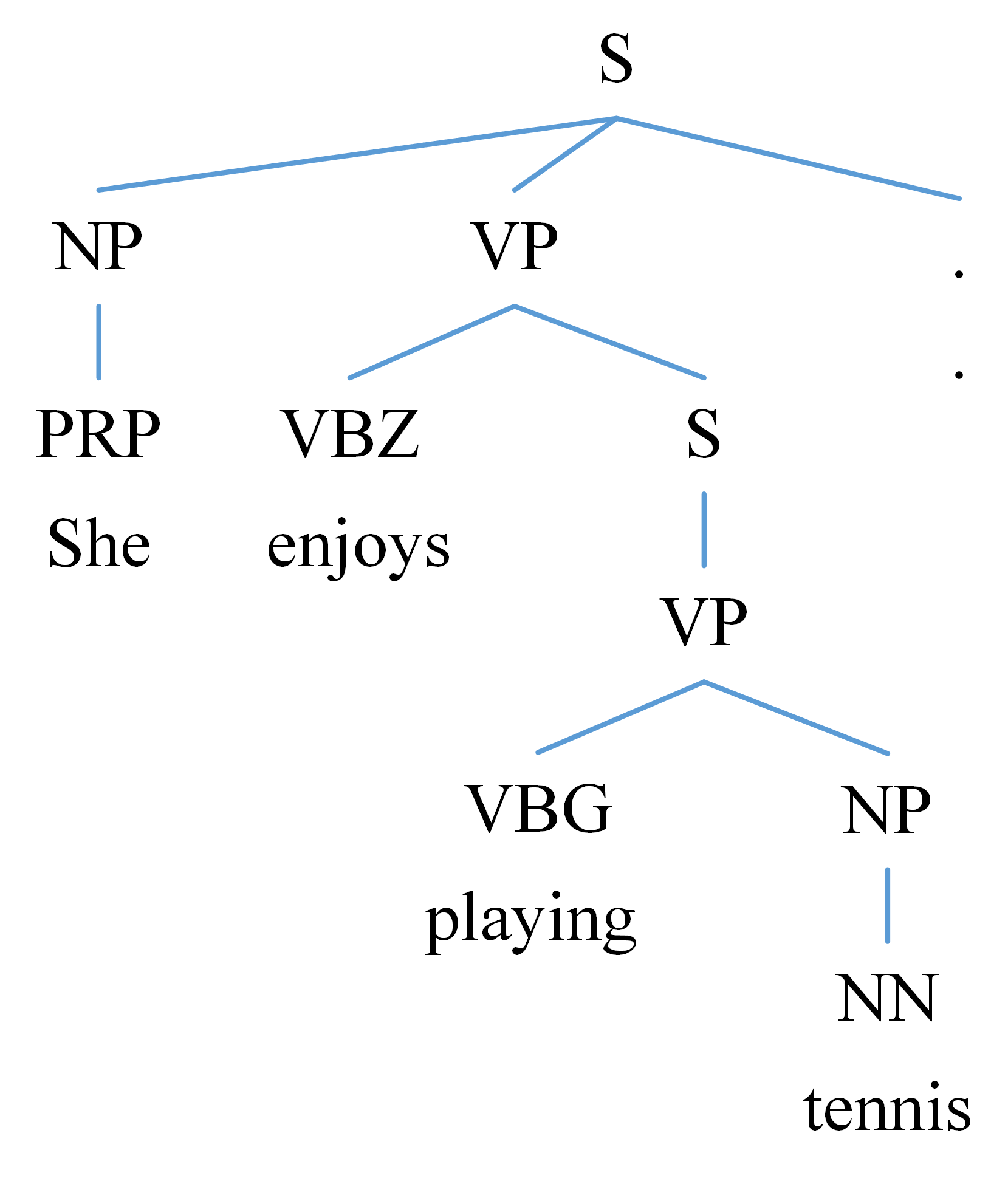}
        \caption{The corresponding inferred grammar tree.}
        \label{fig_tree}
    \end{subfigure}
    \caption{Inferring the parse tree with Algorithm~\ref{alg:distance2tree} given distances, constituent labels, and POS tags. Starting with the full sentence, we pick split point 1 (as it is assigned to the larger distance) and assign label S to span (0,5). The left child span (0,1) is assigned with a tag PRP and a label NP, which produces an unary node and a terminal node. The right child span (1,5) is assigned the label $\varnothing$, coming from implicit binarization, which indicates that the span is not a real constituent and all of its children are instead direct children of its parent. For the span (1,5), the split point 4 is selected. The recursion of splitting and labeling continues until the process reaches a terminal node.}
\label{fig_distance_to_tree}
\end{figure*}

In this section, we start from the concept of syntactic distance introduced in \citet{shen2017neural} for unsupervised parsing via language modeling and we extend it to the supervised setting.
We propose two algorithms, one to convert a parse tree into a compact
representation based on distances between consecutive words, and another to map the inferred representation back to a complete parse tree. The representation will later be used for supervised training.
We formally define the syntactic distances of a parse tree as follows:
\begin{definition} \label{def:distance}
Let $\mathbf{T}$ be a parse tree that contains a set of leaves $(w_0, ..., w_n)$. The height of the lowest common ancestor for two leaves $(w_i, w_j)$ is noted as $\tilde{d}^i_j$. The syntactic distances of $\mathbf{T}$ can be any vector of scalars $\mathbf{d} = (d_1,...,d_n)$ that satisfy:
  \begin{equation}
  \mathrm{sign}( d_{ i }-d_{ j } ) = \mathrm{sign}( { \tilde { d }^{ i-1 }_{ i } }-{ \tilde { d } ^{ j-1 }_{ j } } )
  \end{equation}
\end{definition}

In other words, $\mathbf{d}$ induces the same ranking order as the quantities $\tilde{d}_i^j$ computed between pairs of consecutive words in the sequence,~i.e. $(\tilde{d}^0_1,...,\tilde{d}^{n-1}_n)$. Note that there are $n-1$ syntactic distances for a sentence of length $n$.

\begin{exmp} Consider the tree in Fig.~\ref{fig_dis2tree_example} for which $\tilde d_1^0 = 2$, $\tilde d_2^1 = 1$. An example of valid syntactic distances for this tree is any $\mathbf{d} = (d_1, d_2)$ such that $d_1 > d_2$.
\end{exmp}

Given this definition, the parsing model predicts a sequence of scalars, which is a more natural setting for models based on neural networks, rather than predicting a set of spans.
For comparison, in most of the current neural parsing methods, the model needs to output a sequence of transitions~\citep{cross2016span,chen2014fast}.

\begin{algorithm}[t]
  \caption{Binary Parse Tree to Distance}\label{alg:tree2distance}
  ($\cup$ represents the concatenation operator of lists)
  \begin{algorithmic}[1]
    \Function{Distance}{node}
        \If {node \textbf{is} leaf}
            \State {$\mathbf{d} \gets []$}
            \State {$\mathbf{c} \gets []$}
            \State {$\mathbf{t} \gets [\mathrm{node.tag}]$}
            \State {$h \gets 0$}
        \Else
            \State {$\mathrm{child}_l, \mathrm{child}_r \gets$ children of node}
            \State {$\mathbf{d}_l$, $\mathbf{c}_l$, $\mathbf{t}_l$, $h_l \gets$ Distance($\mathrm{child}_l$)}
            \State {$\mathbf{d}_r$, $\mathbf{c}_r$, $\mathbf{t}_r$, $h_r \gets$ Distance($\mathrm{child}_r$)}

            \State {$h \gets \max (h_l,h_r) + 1$}
            \State {$\mathbf{d} \gets \mathbf{d}_l \cup [h] \cup \mathbf{d}_r$}
            \State {$\mathbf{c} \gets \mathbf{c}_l \cup [\mathrm{node.label}] \cup \mathbf{c}_r$}
            \State {$\mathbf{t} \gets \mathbf{t}_l \cup \mathbf{t}_r$}
        \EndIf
        \State \Return $\mathbf{d}$, $\mathbf{c}$, $\mathbf{t}$, $h$
    \EndFunction
  \end{algorithmic}
\end{algorithm}

\begin{algorithm}[t]
\caption{Distance to Binary Parse Tree}\label{alg:distance2tree}
\begin{algorithmic}[1]
\Function{Tree}{$\mathbf{d}$,$\mathbf{c}$,$\mathbf{t}$}
	\If {$\mathbf{d} = []$}
    	\State {node $\gets$ Leaf($\mathbf{t}$)}
    \Else
    	\State {$i \gets \mathrm{arg}\max_i (\mathbf{d})$}
        \State {$\mathrm{child}_l$ $\gets$ Tree($\mathbf{d}_{<i}$, $\mathbf{c}_{<i}$, $\mathbf{t}_{< i}$)}
        \State {$\mathrm{child}_r$ $\gets$ Tree($\mathbf{d}_{>i}$, $\mathbf{c}_{>i}$, $\mathbf{t}_{\geq i}$)}
        \State $\mathrm{node} \gets$ Node($\mathrm{child}_l$, $\mathrm{child}_r$, $\mathbf{c}_i$)
    \EndIf
    \State \Return $\mathrm{node}$
\EndFunction
\end{algorithmic}
\end{algorithm}

Let us first consider the case of a binary parse tree. Algorithm~\ref{alg:tree2distance} provides a way to convert it to a tuple $(\mathbf{d},\mathbf{c},\mathbf{t})$, where $\mathbf{d}$ contains the height of the inner nodes in the tree following a left-to-right (in order) traversal, $\mathbf{c}$ the constituent labels for each node in the same order and $\mathbf{t}$ the part-of-speech (POS) tags of each word in the left-to-right order. $\mathbf{d}$ is a valid vector of syntactic distances satisfying Definition \ref{def:distance}.

Once a model has learned to predict these variables, Algorithm \ref{alg:distance2tree} can reconstruct a unique binary tree from the output of the model $(\mathbf{\hat d},\mathbf{\hat c},\mathbf{\hat t})$. The idea in Algorithm~\ref{alg:distance2tree} is similar to the top-down parsing method proposed by~\citet{stern2017minimal}, but differs in one key aspect: at each recursive call, there is no need to estimate the confidence for every split point. The algorithm simply chooses the split point $i$ with the maximum $\hat d_i$, and assigns to the span the predicted label $\hat c_i$. This makes the running time of our algorithm to be in $\mathcal{O}(n\log n)$, compared to the $\mathcal{O}(n^2)$ of the greedy top-down algorithm by~\citep{stern2017minimal}. Figure~\ref{fig_distance_to_tree} shows an example of the reconstruction of parse tree.
Alternatively, the tree reconstruction process can also be done in a bottom-up manner, which requires the recursive composition of adjacent spans according to the ranking induced by their syntactic distance, a process akin to agglomerative clustering.

One potential issue is the existence of unary and $n$-ary nodes. 
We follow the method proposed by~\citet{stern2017minimal} and add a special empty label $\varnothing$ to spans that are not themselves full constituents but simply arise during the course of implicit binarization.
For the unary nodes that contains one nonterminal node, we take the common approach of treating these as additional atomic labels alongside all elementary nonterminals~\citep{stern2017minimal}.
For all terminal nodes, we determine whether it belongs to a unary chain or not by predicting an additional label. If it is predicted with a label different from the empty label, we conclude that it is a direct child of a unary constituent with that label. Otherwise if it is predicted to have an empty label, we conclude that it is a child of a bigger constituent which has other constituents or words as its siblings.

An $n$-ary node can arbitrarily be split into binary nodes. We choose to use the leftmost split point. The split point may also be chosen based on model prediction during training. Recovering an $n$-ary parse tree from the predicted binary tree simply requires removing the empty nodes and split combined labels corresponding to unary chains. 

Algorithm \ref{alg:distance2tree} is a divide-and-conquer algorithm. The running time of this procedure is $\mathcal{O}(n\log n)$. However, the algorithm is naturally adapted for execution in a parallel environment, which can further reduce its running time to $\mathcal{O}(\log n)$.

\section{Learning Syntactic Distances}
We use neural networks to estimate the vector of syntactic distances for a given sentence. We use a modified hinge loss, where the target distances are generated by the tree-to-distance conversion given by Algorithm \ref{alg:tree2distance}. Section \ref{model_architecture} will describe in detail the model architecture, and Section \ref{objective} describes the loss we use in this setting.




\begin{figure*}[!h]
	\centering
    \includegraphics[width=0.8\textwidth]{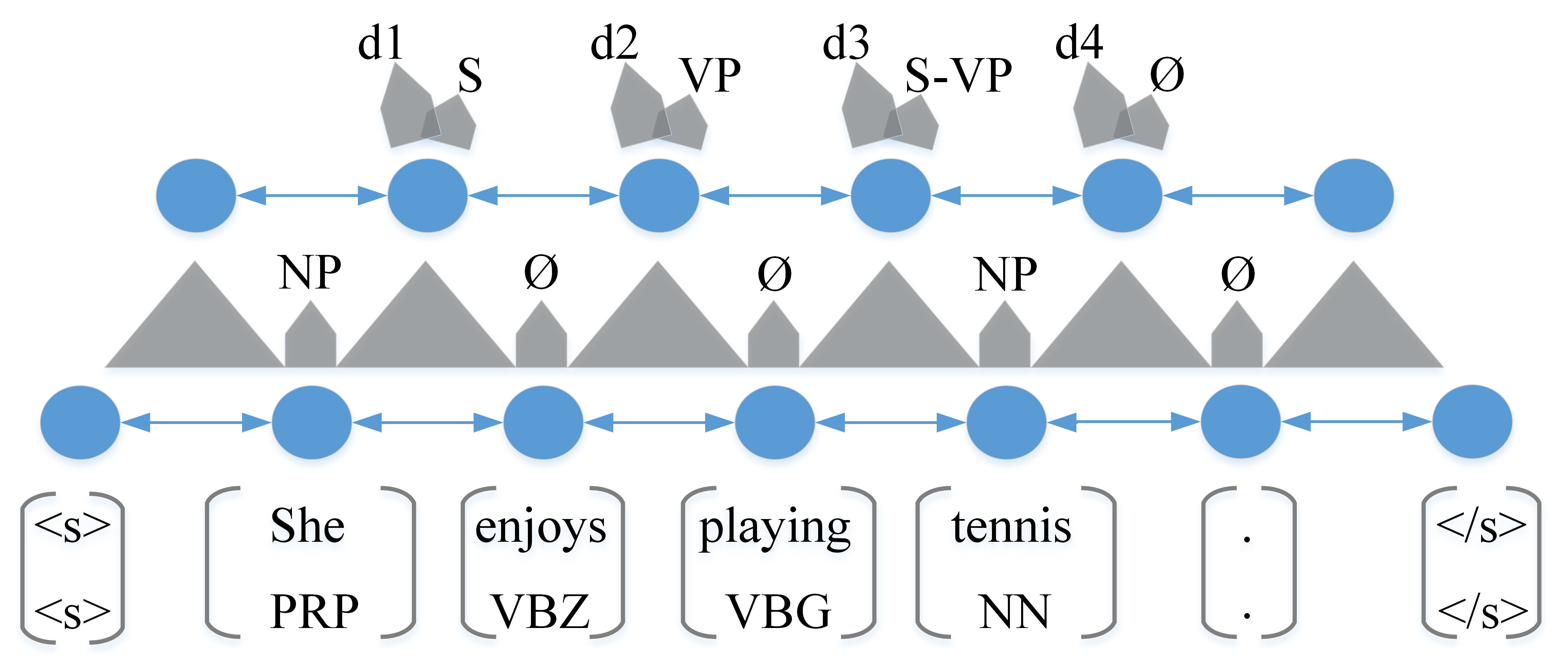}
    \caption{The overall visualization of our model. 
    Circles represent hidden states, triangles represent convolution layers, block arrows represent feed-forward layers, arrows represent recurrent connections.
    The bottom part of the model predicts unary labels for each input word.
    The $\emptyset$ is treated as a special label together with other labels.
    The top part of the model predicts the syntactic distances and the constituent labels. 
    The inputs of model are the word embeddings concatenated with the POS tag embeddings. The tags are given by an external Part-Of-Speech tagger.
    }\label{fig_model}
\end{figure*}

\subsection{Model Architecture}	\label{model_architecture}
Given input words $\mathbf{w} = (w_0, w_1, ..., w_n)$, we predict the tuple $(\mathbf{d}, \mathbf{c}, \mathbf{t})$. 
The POS tags $\mathbf{t}$ are given by an external Part-Of-Speech (POS) tagger. The syntactic distances $\mathbf{d}$ and constituent labels $\mathbf{c}$ are predicted using a neural network architecture that stacks recurrent (LSTM~\citep{hochreiter1997long}) and convolutional layers. 

Words and tags are first mapped to sequences of embeddings $\mathbf{e}_0^\mathrm{w}, ..., \mathbf{e}_n^\mathrm{w}$ and $\mathbf{e}_0^\mathrm{t}, ..., \mathbf{e}_n^\mathrm{t}$.
Then the word embeddings and the tag embeddings are concatenated together as inputs for a stack of bidirectional LSTM layers:
\begin{equation}
\mathbf{h}_0^\mathrm{w}, ..., \mathbf{h}_n^\mathrm{w} = \text{BiLSTM}_\mathrm{w}([\mathbf{e}_0^\mathrm{w}, \mathbf{e}_0^\mathrm{t}], ..., [\mathbf{e}_n^\mathrm{w}, \mathbf{e}_n^\mathrm{t}])
\end{equation}
where $\text{BiLSTM}_\mathrm{w}(\cdot)$ is the word-level bidirectional layer, which gives the model enough capacity to capture long-term syntactical relations between words. 

To predict the constituent labels for each word, we pass the hidden states representations $\mathbf{h}_0^\mathrm{w}, ..., \mathbf{h}_n^\mathrm{w}$ through a 2-layer network $\text{FF}^\mathrm{w}_c$, with softmax output:
\begin{equation}
p(c_i^\mathbf{w} | \mathbf{w}) = \text{softmax}(\text{FF}^\mathrm{w}_c(\mathbf{h}_i^\mathrm{w})) \label{eq:out_lab_terminal}
\end{equation}


To compose the necessary information for inferring the syntactic distances and the constituency label information, we perform an additional convolution:
\begin{equation}
\mathbf{g}_1^\mathrm{s}, \ldots, \mathbf{g}_n^\mathrm{s} = \text{CONV}(\mathbf{h}_0^\mathrm{w}, ..., \mathbf{h}_n^\mathrm{w})
\end{equation}
where $\mathbf{g}_i^\mathrm{s}$ can be seen as a draft representation for each split position in Algorithm \ref{alg:distance2tree}. Note that the subscripts of $g_i^s$s start with 1, since we have $n-1$ positions as non-terminal constituents. Then, we stack a bidirectional LSTM layer on top of $\mathbf{g}_i^\mathrm{s}$:
\begin{eqnarray}
\mathbf{h}_1^\mathrm{s}, ..., \mathbf{h}_n^\mathrm{s} = \text{BiLSTM}_\mathrm{s}(\mathbf{g}_1^\mathrm{s}, \ldots, \mathbf{g}_n^\mathrm{s})
\end{eqnarray}
where $\text{BiLSTM}_\mathrm{s}$ fine-tunes the representation by conditioning on other split position representations.
Interleaving between LSTM and convolution layers turned out empirically to be the best choice over multiple variations of the model, including using self-attention~\citep{vaswani2017attention} instead of LSTM.

To calculate the syntactic distances for each position, the vectors $\mathbf{h}_1^\mathrm{s}, \ldots, \mathbf{h}_n^\mathrm{s}$ are transformed through a 2-layer feed-forward network $\text{FF}_d$ with a single output unit (this can be done in parallel with 1x1 convolutions), with no activation function at the output layer:
\begin{equation}
\label{eq:out_dst}
\hat d_i = \text{FF}_d(\mathbf{h}_i^s),
\end{equation}
For predicting the constituent labels, we pass the same representations $\mathbf{h}_1^\mathrm{s}, \ldots, \mathbf{h}_n^\mathrm{s}$ through another 2-layer network $\text{FF}_c^\mathrm{s}$, with softmax output. 
\begin{equation}
p(c_i^\mathrm{s} | \mathbf{w}) = \text{softmax}(\text{FF}^\mathrm{s}_c(\mathbf{h}_i^\mathrm{s})) \label{eq:out_lab_nonterminal}
\end{equation}

The overall architecture is shown in Figure \ref{fig_infer_tree}. Since the output $( \mathbf{d}, \mathbf{c}, \mathbf{t} )$ can be unambiguously transfered to a unique parse tree, the model implicitly makes all parsing decisions inside the recurrent and convolutional layers.


\subsection{Objective}	\label{objective}
Given a set of training examples $\mathcal{D} = \{\langle \mathbf{d}_k, \mathbf{c}_k, \mathbf{t}_k, \mathbf{w}_k\rangle\}_{k=1}^{K}$, the training objective is the sum of the prediction losses of syntactic distances $\mathbf{d}_k$ and constituent labels $\mathbf{c}_k$. 

Due to the categorical nature of variable $\mathbf{c}$, we use a standard softmax classifier with a cross-entropy loss $L_{\text{label}}$ for constituent labels, using the estimated probabilities obtained in Eq.~\ref{eq:out_lab_terminal} and \ref{eq:out_lab_nonterminal}.

A na\"ive loss function for estimating syntactic distances is the mean-squared error (MSE):
\begin{equation}
L^{\text{mse}}_{\text{dist}}  = \sum_i (d_{i} - \hat d_{i})^{2}
\end{equation}
The MSE loss forces the model to regress on the exact value of the true distances. Given that only the \emph{ranking} induced by the ground-truth distances in $\mathbf{d}$ is important, as opposed to the absolute values themselves, using an MSE loss over-penalizes the model by ignoring ranking equivalence between different predictions.

Therefore, we propose to minimize a pair-wise learning-to-rank loss, similar to those proposed in \citep{RankNet}. We define our loss as a variant of the hinge loss as:
\begin{equation}
L^{\text{rank}}_{\text{dist}} = \sum_{i, j > i} [1 - \text{sign}(d_i - d_j) (\hat d_i - \hat d_j)]^+,
\end{equation}
where $[x]^+$ is defined as $\max(0, x)$. This loss encourages the model to reproduce the full ranking order induced by the ground-truth distances. The final loss for the overall model is just the sum of individual losses $L = L_{\text{label}} + L^\text{rank}_{\text{dist}}$.


\section{Experiments}

\begin{table}[t]  
\centering  
  \begin{tabular}{l c c c}
    \hline
    Model & LP & LR & F1 \\
    \hline
    \textbf{Single Model} \\
    \hline
    \citet{vinyals2015grammar} 					& - & - & 88.3 \\
    \citet{zhu2013fast}        					& 90.7 & 90.2 & 90.4 \\
   	\citet{dyer2016recurrent}					& - & - & 89.8 \\
	\citet{watanabe2015transition} 				& - & - & 90.7 \\
	\citet{cross2016span}	 					& 92.1 & 90.5 & 91.3 \\
    \citet{liu2016shift}						& 92.1 & 91.3 & 91.7 \\
    \citet{stern2017minimal}					& 93.2 & 90.3 & 91.8 \\
    \citet{liu2017order}						& - & - & 91.8 \\
    \citet{gaddy2018what} 						& - & - & 92.1 \\
    \citet{stern2017effective} 					& 92.5 & 92.5 & 92.5 \\
    \hline
    \textbf{Our Model}							& 92.0 & 91.7 & 91.8\\
    \hline
    \hline
    \textbf{Ensemble} \\
    \hline
    \citet{shindo2012bayesian}					& - & - & 92.4 \\
    \citet{vinyals2015grammar}					& - & - & 90.5 \\
    \hline
	\textbf{Semi-supervised} \\
    \hline
    \citet{zhu2013fast} 						& 91.5 & 91.1 & 91.3 \\
    \citet{vinyals2015grammar}					& - & - & 92.8 \\
    \hline
    \textbf{Re-ranking} \\
    \hline
    \citet{charniak2005coarse}					& 91.8 & 91.2 & 91.5 \\
    \citet{huang2008forest}						& 91.2 & 92.2 & 91.7 \\
    \citet{dyer2016recurrent}					& - & - & 93.3 \\
    \hline
  \end{tabular}
  \caption{Results on the PTB dataset WSJ test set, Section 23. LP, LR represents labeled precision and recall respectively.}
  \label{table_ptb}
\end{table}

We evaluate our model described above on 2 different datasets, the standard Wall Street Journal (WSJ) part of the Penn Treebank (PTB) dataset, and the Chinese Treebank (CTB) dataset. 

For evaluating the F1 score, we use the standard \texttt{evalb}\footnote{\url{http://nlp.cs.nyu.edu/evalb/}} tool. We provide both labeled and unlabeled F1 score, where the former takes into consideration the constituent label for each predicted constituent, while the latter only considers the position of the constituents. In the tables below, we report the labeled F1 scores for comparison with previous work, as this is the standard metric usually reported in the relevant literature.

\subsection{Penn Treebank}

For the PTB experiments, we follow the standard train/valid/test separation and use sections 2-21 for training, section 22 for development and section 23 for test set. Following this split, the dataset has 45K training sentences and 1700, 2416 sentences for valid/test respectively. The placeholders with the \texttt{-NONE-} tag are stripped from the dataset during preprocessing. The POS tags are predicted with the Stanford Tagger \citep{toutanova2003feature}. 

We use a hidden size of 1200 for each direction on all LSTMs, with 0.3 dropout in all the feed-forward connections, and 0.2 recurrent connection dropout \citep{merity2017regularizing}. The convolutional filter size is 2. The number of convolutional channels is 1200. As a common practice for neural network based NLP models, the embedding layer that maps word indexes to word embeddings is randomly initialized. The word embeddings are sized 400. Following~\cite{merity2017regularizing}, we randomly swap an input word embedding during training with the zero vector with probability of 0.1. We found this helped the model to generalize better. Training is conducted with Adam algorithm with l2 regularization decay $1\times10^{-6}$. We pick the result obtaining the highest labeled F1 on the validation set, and report the corresponding test F1, together with other statistics. We report our results in Table~\ref{table_ptb}. Our best model obtains a labeled F1 score of 91.8 on the test set (Table \ref{table_ptb}). Detailed dev/test set performances, including label accuracy is reported in Table \ref{table_ptbctb}.

Our model performs achieves good performance for single-model constituency parsing trained without external data. The best result from~\citep{stern2017effective} is obtained by a generative model. Very recently, we came to knowledge of~\citet{gaddy2018what}, which uses character-level LSTM features coupled with chart-based parsing to improve performance. Similar sub-word features can be also used in our model. We leave this investigation for future works.
For comparison, other models obtaining better scores either use ensembles, benefit from semi-supervised learning, or recur to re-ranking of a set of candidates.

\subsection{Chinese Treebank}
We use the Chinese Treebank 5.1 dataset, with articles 001-270 and 440-1151 for training, articles 301-325 as development set, and articles 271-300 for test set. This is a standard split in the literature \cite{liu2016shift}. The \texttt{-NONE-} tags are stripped as well. The hidden size for the LSTM networks is set to 1200. We use a dropout rate of 0.4 on the feed-forward connections, and 0.1 recurrent connection dropout. The convolutional layer has 1200 channels, with a filter size of 2. We use 400 dimensional word embeddings. During training, input word embeddings are randomly swapped with the zero vector with probability of 0.1. We also apply a l2 regularization weighted by $1\times10^{-6}$ on the parameters of the network. Table~\ref{table_ctb} reports our results compared to other benchmarks. To the best of our knowledge, we set a new state-of-the-art for single-model parsing achieving 86.5 F1 on the test set. The detailed statistics are shown in Table~\ref{table_ptbctb}.

\begin{table}[t]  
\centering  
  \begin{tabular}{l c c c}
    \hline
    Model & LP & LR & F1 \\
    \hline
    \textbf{Single Model} \\
    \hline
    \citet{charniak2000maximum}		& 82.1 & 79.6 & 80.8 \\
    \citet{zhu2013fast} 			& 84.3 & 82.1 & 83.2 \\
    \citet{wang2015feature}			& - & - & 83.2 \\
    \citet{watanabe2015transition}	& - & - & 84.3 \\
    \citet{dyer2016recurrent}		& - & - & 84.6 \\
    \citet{liu2016shift}			& 85.9 & 85.2 & 85.5 \\
    \citet{liu2017order} 			& - & - & 86.1 \\
    \hline
    \textbf{Our Model}				& 86.6 & 86.4 & 86.5\\
    \hline
    \hline
    \textbf{Semi-supervised} \\
    \hline
    \citet{zhu2013fast} 			& 86.8 & 84.4 & 85.6 \\
    \citet{wang2014joint}			& -    & -    & 86.3 \\
    \citet{wang2015feature}			& -    & -    & 86.6 \\
    \hline
    \textbf{Re-ranking} \\
    \hline
	\citet{charniak2005coarse}		& 83.8 & 80.8 & 82.3 \\
    \citet{dyer2016recurrent}		& - & - & 86.9 \\
    \hline
  \end{tabular}
  \caption{Test set performance comparison on the CTB dataset}
  \label{table_ctb}
\end{table}

\begin{table*}[!ht]
\centering  
  \begin{tabular}{ c | c | c c c | c }
    \hline
    \multicolumn{2}{c | }{dev/test result} & Prec. & Recall & F1 & label accuracy \\
    \hline
    \multirow{2}{*}{PTB} & labeled		& 91.7/92.0		& 91.8/91.7		& 91.8/91.8		& \multirow{2}{*}{94.9/95.4\%} \\
    \cline{2-5}
    					 & unlabeled	& 93.0/93.2		& 93.0/92.8		& 93.0/93.0		& 			\\                         
    \hline
    \multirow{2}{*}{CTB} & labeled		& 89.4/86.6		& 89.4/86.4		& 89.4/86.5		& \multirow{2}{*}{92.2/91.1\%} \\
    \cline{2-5}
    					 & unlabeled	& 91.1/88.9		& 91.1/88.6		& 91.1/88.8		& 			\\
    \hline
  \end{tabular}
  \caption{Detailed experimental results on PTB and CTB datasets}
  \label{table_ptbctb}
\end{table*}

\subsection{Ablation Study}
\begin{table}[!ht]
\centering  
  \begin{tabular}{ l c c c }
    \hline
    Model & LP & LR & F1 \\
    \hline
    Full model 			& 92.0 & 91.7 & 91.8 \\
    \hline
    \hline
    w/o top LSTM			& 91.0 & 90.5 & 90.7 \\
    w. Char LSTM & 92.1 & 91.7 & 91.9 \\
    w. embedding 			& 91.9 & 91.6 & 91.7 \\
    w. MSE loss 		  	& 90.3 & 90.0 & 90.1 \\
    \hline
  \end{tabular}
  \caption{Ablation test on the PTB dataset. 
  ``w/o top LSTM'' is the full model without the top LSTM layer. 
  ``w Char LSTM'' is the full model with the extra Character-level LSTM layer. 
  ``w. embedding'' stands for the full model using the pretrained word embeddings. 
  ``w. MSE loss'' stands for the full model trained with MSE loss.}
  \label{table_ablation}
\end{table}

We perform an ablation study by removing/adding components from a our model, and re-train the ablated version from scratch. This gives an idea of the relative contributions of each of the components in the model. Results are reported in Table~\ref{table_ablation}. It seems that the top LSTM layer has a relatively big impact on performance. This may give additional capacity to the model for capturing long-term dependencies useful for label prediction. We used an extra 1-layer character-level BiLSTM to compute an extra word level embedding vector as input of our model. It's seems that character-level features give marginal improvements in our model. We also experimented by using 300D GloVe~\citep{pennington2014glove} embedding for the input layer but this didn't yield improvements over the model's best performance. Unsurprisingly, the model trained with MSE loss underperforms considerably a model trained with the rank loss. 


\subsection{Parsing Speed}
The prediction of syntactic distances can be batched in modern GPU architectures. The distance to tree conversion is a $\mathcal{O}(n \log n)$ ($n$ stand for the number of words in the input sentence) divide-and-conquer algorithm.
We compare the parsing speed of our parser with other state-of-the-art neural parsers in Table~\ref{table_speed}. As the syntactic distance computation can be performed in parallel within a GPU, we first compute the distances in a batch, then we iteratively decode the tree with Algorithm~\ref{alg:distance2tree}.
It is worth to note that this comparison may be unfair since some of the reported results may use very different hardware settings. We couldn't find the source code to re-run them on our hardware, to give a fair enough comparison.
In our setting, we use an NVIDIA TITAN Xp graphics card for running the neural network part, and the distance to tree inference is run on an Intel Core i7-6850K CPU, with 3.60GHz clock speed.

\begin{table}[!h]  
\centering  
  \begin{tabular}{l|c}
    \hline
    Model & \# sents/sec\\
    \hline
    \citet{petrov2007improved}		& 6.2 \\
    \citet{zhu2013fast}          	& 89.5 \\
    \citet{liu2016shift}			& 79.2 \\
    \citet{stern2017minimal} 		& 75.5 \\
    \hline
    \hline
    Our model & 111.1 \\
    Our model w/o tree inference & 351 \\
    \hline
  \end{tabular}
  \caption{Parsing speed in sentences per second on the PTB dataset.}
  \label{table_speed}
\end{table}

\section{Related Work}
Parsing natural language with neural network models has recently received growing attention. These models have attained state-of-the-art results for dependency parsing~\citep{chen2014fast} and constituency parsing~\citep{dyer2016recurrent,cross2016span, coavoux2016neural}. Early work in neural network based parsing directly use a feed-forward neural network to predict parse trees~\citep{chen2014fast}.~\citet{vinyals2015grammar} use a sequence-to-sequence framework where the decoder outputs a linearized version of the parse tree given an input sentence. Generally, in these models, the correctness of the output tree is not strictly ensured (although empirically observed).

Other parsing methods ensure structural consistency by operating in a transition-based setting~\citep{chen2014fast} by parsing either in the top-down direction~\citep{dyer2016recurrent,liu2016shift}, bottom-up~\citep{zhu2013fast,watanabe2015transition,cross2016span} and recently in-order~\citep{liu2017order}. Transition-based methods generally suffer from compounding errors due to exposure bias: during testing, the model is exposed to a very different regime (i.e.~decisions sampled from the model itself) than what was encountered during training (i.e.~the ground-truth decisions)~\citep{daume2009search,goldberg2012dynamic}. This can have catastrophic effects on test performance but can be mitigated to a certain extent by using beam-search instead of greedy decoding. \cite{stern2017effective} proposes an effective inference method for generative parsing, which enables direct decoding in those models. More complex training methods have been devised in order to alleviate this problem~\citep{goldberg2012dynamic,cross2016span}. Other efforts have been put into neural chart-based parsing~\citep{durrett2015neural,stern2017minimal} which ensure structural consistency and offer exact inference with CYK algorithm. \cite{gaddy2018what} includes a simplified CYK-style inference, but the complexity still remains in $O(n^3)$. 

In this work, our model learns to produce a particular representation of a tree in parallel. Representations can be computed in parallel, and the conversion from representation to a full tree can efficiently be done with a divide-and-conquer algorithm. As our model outputs decisions in parallel, our model doesn't suffer from the exposure bias. Interestingly, a series of recent works, both in machine translation~\citep{gu2017non} and speech synthesis~\citep{oord2017parallel}, considered the sequence of output variables conditionally independent given the inputs.

\section{Conclusion}
We presented a novel constituency parsing scheme based on predicting real-valued scalars, named syntactic distances, whose ordering identify the sequence of top-down split decisions. We employ a neural network model that predicts the distances $\mathbf{d}$ and the constituent labels $\mathbf{c}$. Given the algorithms presented in Section \ref{sec:dis_n_tree}, we can build an unambiguous mapping between each $(\mathbf{d}, \mathbf{c}, \mathbf{t})$ and a parse tree. One peculiar aspect of our model is that it predicts split decisions~\emph{in parallel}. Our experiments show that our model can achieve strong performance compare to previous models, while being significantly more efficient. Since the architecture of model is no more than a stack of standard recurrent and convolution layers, which are essential components in most academic and industrial deep learning frameworks, the deployment of this method would be straightforward.

\section*{Acknowledgement}
The authors would like to thank Compute Canada for providing the computational resources. The authors would also like to thank Jackie Chi Kit Cheung for the helpful discussions. Zhouhan Lin would like to thank AdeptMind for generously supporting his research via scholarship. 

\bibliography{acl2018}
\bibliographystyle{acl_natbib}
\end{document}